\documentclass{article}

\usepackage{arxiv}

\usepackage[utf8]{inputenc} 
\usepackage[T1]{fontenc}    
\usepackage{hyperref}       
\usepackage{url}            
\usepackage{booktabs}       
\usepackage{amsfonts}       
\usepackage{nicefrac}       
\usepackage{microtype}      
\usepackage{lipsum}
\usepackage{graphicx}
\usepackage{natbib}
\usepackage{amsmath}
\usepackage{natbib}
\usepackage{bibentry}
\usepackage{multirow}
\usepackage{caption}
\usepackage{subcaption}
\graphicspath{ {./images/} }

\newcommand{\algopt}{G$^2$RPO-A\xspace}
\usepackage{xspace}

\title{G$^2$RPO-A: Guided Group Relative Policy Optimization with Adaptive Guidance}

\author{Yongxin Guo$^{1,2}$\thanks{Equal Contribution.}, Wenbo Deng$^{1*}$, Zhenglin Cheng$^{3}$, Xiaoying Tang$^{1}$ \\
$^1$School of Science and Engineering, The Chinese University of Hong Kong, Shenzhen, \\
Guangdong, 518172, P.R. China
$^2$Alibaba Group \\
$^3$School of Engineering, Westlake University \\
}

\begin{document}
\maketitle
\begin{abstract}
Reinforcement Learning with Verifiable Rewards (RLVR) has markedly enhanced the reasoning abilities of large language models (LLMs). Its success, however, largely depends on strong base models with rich world knowledge, yielding only modest improvements for small-size language models (SLMs).  
To address this limitation, we investigate Guided GRPO, which injects ground-truth reasoning steps into roll-out trajectories to compensate for SLMs’ inherent weaknesses. Through a comprehensive study of various guidance configurations, we find that naively adding guidance delivers limited gains. These insights motivate \algopt, an adaptive algorithm that automatically adjusts guidance strength in response to the model’s evolving training dynamics. 
Experiments on mathematical reasoning and code-generation benchmarks confirm that \algopt substantially outperforms vanilla GRPO. Our code and models are available at \url{https://github.com/T-Lab-CUHKSZ/G2RPO-A}.
\end{abstract}


\section{Introduction}
Recent advancements in reasoning-centric large language models (LLMs), exemplified by DeepSeek-R1 \cite{guo2025deepseek}, OpenAI-o1 \cite{jaech2024openai}, and Qwen3 \cite{yang2025qwen3}, have significantly expanded the performance boundaries of LLMs, showcasing the immense potential of reasoning-enhanced models.
Building upon robust base models with comprehensive world knowledge, these reasoning-focused LLMs have achieved breakthrough progress in complex domains such as mathematics \cite{guan2025rstarmath}, coding \cite{souza2025codegenerationsmalllanguage,huang2025efficoder}, and other grounding tasks \citep{li2025torl,wei2025swe}. 
At the core of this success lies Reinforcement Learning with Verifiable Rewards (RLVR) \citep{shao2024deepseekmathpushinglimitsmathematical,chu2025sft,liu2025understandingr1zeroliketrainingcritical}. This innovative approach, which employs reinforcement learning techniques in LLMs using rule-based outcome rewards, has garnered significant attention in the AI community. RLVR has demonstrated remarkable improvement in generalization across a wide spectrum of downstream tasks \citep{jia2025writingzerobridgegapnonverifiable,wu2025rlvrworldtrainingworldmodels}, positioning it as a pivotal advancement in the field of artificial intelligence.

As the de-facto algorithm, Group Relative Policy Optimization (GRPO)~\citep{shao2024deepseekmathpushinglimitsmathematical} improves upon Proximal Policy Optimization (PPO)~\citep{schulman2017proximalpolicyoptimizationalgorithms} by removing the need for a critic model through inner-group response comparison, thereby speeding up the training.

\begin{figure}[t]
    \centering
    \includegraphics[width=0.5\columnwidth]{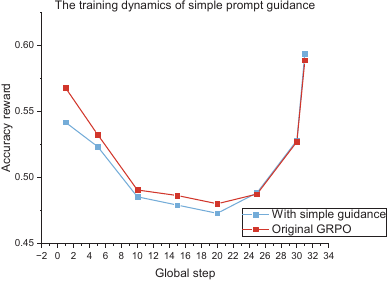}
    \caption{\textbf{Naive guidance does not help.} Using Qwen2.5-Math-7B as the base model, we train it on the s1K-1.1 dataset for a single epoch with a simple, fixed-length guidance (naive guidance). The naive guidance method shows a temporary increase in the accuracy reward during the early training stages, but it quickly becomes indistinguishable from the vanilla GRPO curve.}
    \label{fig:4}
\end{figure}

\begin{figure*}[t]
    \centering
    \includegraphics[width=0.9\textwidth]{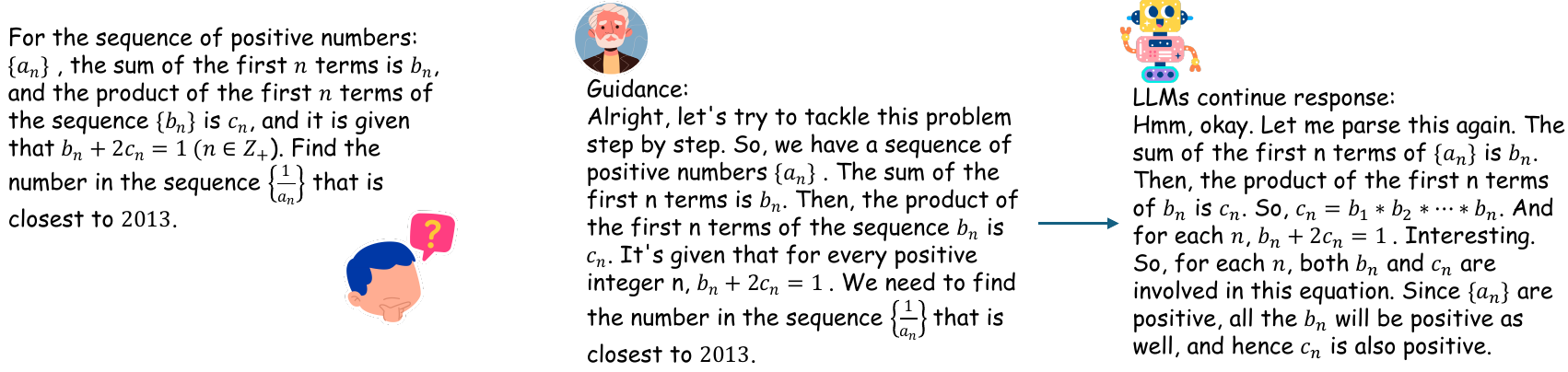}
    \caption{\textbf{Illustration of roll-outs with guidance.} An example of using high‐quality thinking trajectories to guide models.}
    \label{fig:1}
\end{figure*}

\paragraph{Capacity of small-size LLMs limit the performance gains of GRPO.} Despite GRPO's success with large-scale LLMs, its effectiveness is significantly constrained when applied to smaller LLMs. Recent research \citep{ye2025limoreasoning,muennighoff2025s} reveals that GRPO's performance gains highly depend on the base model's capacity \cite{bae2025onlinedifficultyfilteringreasoning, xu2025phi4minireasoningexploringlimitssmall,zhuang2025technicalstudy05breasoning}. Consequently, small-scale LLMs (SLMs) show limited improvement under GRPO (Table ~\ref{tab:benchmark-results-math},~\ref{tab:adagui-ratio-code}), exposing a critical scalability challenge in enhancing reasoning capabilities across diverse model sizes.
To address this challenge, researchers have explored various approaches: distillation \cite{guo2025deepseek}, multi-stage training \cite{xu2025phi4minireasoningexploringlimitssmall} prior to RLVR, and selective sample filtering \cite{xiong2025minimalistapproachllmreasoning,shi2025efficientreinforcementfinetuningadaptive}. However, these methods precede RLVR or suffer performance degradation in complex problems (Table~\ref{tab:dataset-arrangement-results}). Consequently, optimizing the RLVR process for efficient learning in SLMs remains an open challenge, representing a critical frontier in AI research.

\paragraph{Adaptive guidance as a solution.} We propose incorporating guidance into the roll-out process to facilitate the generation of high-quality, reward-worthy candidates (Figure~\ref{fig:1}). However, our initial findings revealed that the implementation of simple fixed-length guidance to the prompts (naive guidance) failed to improve overall performance (Figure \ref{fig:4}). 
Through a comprehensive analysis of the guidance mechanism—varying both the proportion of guided roll-outs within GRPO batches and the guidance length over training epochs—we obtained two key findings:
(1) Code-generation tasks benefit from a higher guidance ratio than mathematical reasoning tasks, and smaller models likewise require more guidance than larger ones.
(2) The optimal guidance length evolves throughout training and is highly context-dependent, rendering simple, predefined schedules ineffective.
In response, we introduce the Guided Group Relative Policy Optimization  with Adaptive Guidance (\algopt) algorithm. This innovative approach dynamically adjusts guidance length based on the model's real-time learning state, offering a sophisticated solution to the challenges of enhancing small-size LLMs' performance in RLVR processes. The key contributions of this paper are summarized as follows:
\begin{itemize}
    \item We enhance GRPO for small-scale LLMs by injecting guidance into the rollout thinking trajectories and conduct a systematic analysis of the effects of key guidance configurations, specifically focusing on the guidance ratio and guidance length.
    \item Our study also examines the importance of hard training samples. We find that integrating these samples into the dataset using a curriculum learning approach, and aided by the guidance mechanism, significantly boosts the training efficiency of our method for SLMs.
    \item Drawing on these findings, we introduce \algopt, an adaptive algorithm that automatically adjusts guidance length in response to the evolving training state. Our experimental results demonstrate the effectiveness of the proposed G$^2$RPO-A algorithm.
    \item We evaluate our method on mathematical reasoning and coding tasks with several models--including the Qwen3 series, DeepSeek-Math-7B-Base, and DeepSeek-Coder-6.7B-Base--and observe substantial performance gains over both vanilla GRPO and simple guided baselines.
\end{itemize}

\section{Related Works}

\begin{figure*}[t]
    \centering
    \includegraphics[width=0.8\textwidth]{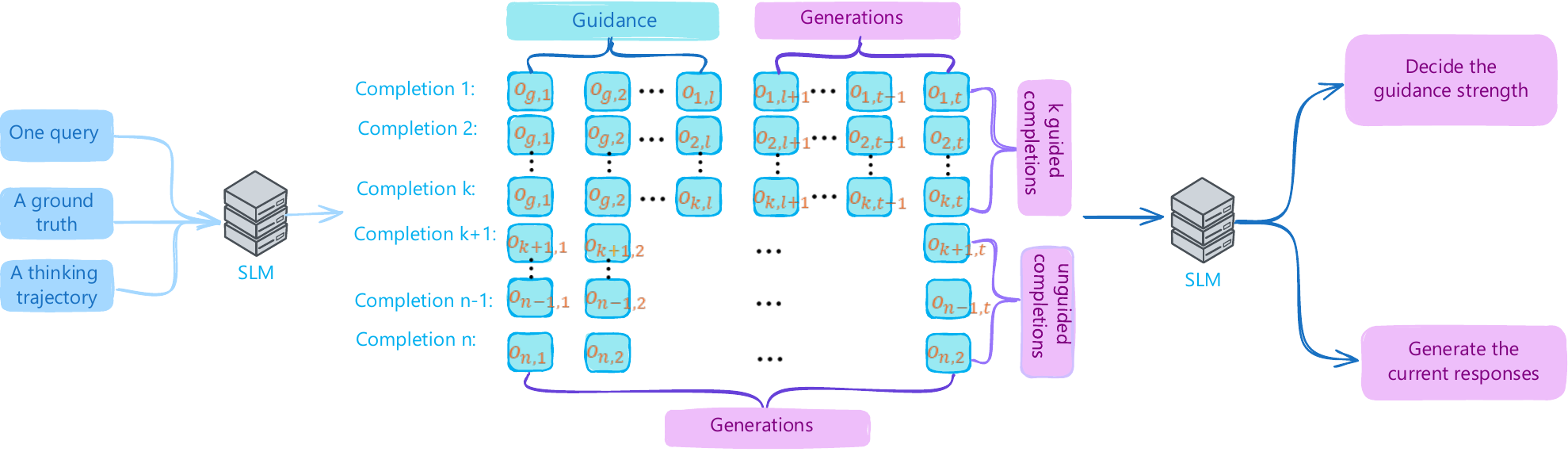}
    \caption{\textbf{Overview of \algopt.}
Each step we split roll-outs into a guided set and an unguided set.
We then compare the current rewards with those from the previous steps; the resulting ratio determines the future guidance length.}
    \label{fig:2}
\end{figure*}

The introduction of chain-of-thought (CoT) prompting has markedly improved LLM performance on complex reasoning tasks \citep{NEURIPS2022_9d560961,kojima2022large}. Complementing this advance, Reinforcement Learning with Verifiable Rewards (RLVR) has emerged as a powerful training paradigm for reasoning-centric language models \cite{yue2025doesreinforcementlearningreally,lee2024token}. The de-facto RLVR algorithm, Group Relative
Policy Optimization (GRPO)~\citep{guo2025deepseek} delivers strong gains on various benchmarks while remaining training-efficient because it dispenses with the need for a separate critic network \cite{wen2025reinforcementlearningverifiablerewards,shao2024deepseekmathpushinglimitsmathematical}. 
Recent efforts to improve GRPO have explored several directions. Some approaches focus on refining the core GRPO objective, either by pruning candidate completions \cite{lin2025cppoacceleratingtraininggroup} or by removing normalization biases \cite{liu2025understandingr1zeroliketrainingcritical}. Separately, another studies aim to enhance the training signal and stability. DAPO~\cite{yu2025dapo}, for instance, introduces dense, step-wise advantage signals and decouples the actor-critic training to mitigate reward sparsity. 

However, adapting GRPO–style algorithms to small-scale LLMs remains challenging due to the sparse-reward problem \cite{lu2024small,vannguyen2024surveysmalllanguagemodels,dang2025reinforcementlearningreasoningsmall}. Recent studies have therefore focused on improved reward estimation \cite{cui2025process}. TreeRPO \cite{yang2025treerpotreerelativepolicy} uses a tree-structured sampling procedure to approximate step-wise expected rewards, and Hint-GRPO \cite{huang2025boostingmllmreasoningtextdebiased} applies several reward-shaping techniques.
Other lines of research investigate knowledge distillation~\citep{guo2025deepseek}, multi-stage pre-training before RLVR \cite{xu2025phi4minireasoningexploringlimitssmall}, and selective sample filtering \cite{xiong2025minimalistapproachllmreasoning,shi2025efficientreinforcementfinetuningadaptive}. In our experiments, however, these filtering or sampling strategies are performed either only before RLVR or do not improve performance on more complex tasks.
In this paper, we introduce a guidance mechanism that injects ground-truth reasoning steps directly into the model’s roll-out trajectories during RL training. Because the guidance is provided online, the proposed method can still learn effectively from difficult examples while mitigating the sparse-reward issue.

The role of guidance in GRPO-style training remains underexplored. Two concurrent studies have addressed related questions \citep{nath2025adaptiveguidanceacceleratesreinforcement,park2025deepvideor1videoreinforcementfinetuning}, but both simply append guidance tokens to the input prompt, offering neither a systematic analysis of guidance configurations nor a mechanism that adapts to the changing training state of the model. We show that naive guidance often fails to improve performance because it yields low expected advantage. To remedy this, we provide a comprehensive examination of how guidance length, ratio, and scheduling affect learning, and we introduce \algopt, an adaptive strategy that dynamically adjusts guidance strength throughout training.

\section{Preliminary}
\paragraph{Group Relative Policy Optimization (GRPO).}

Given a prompt, GRPO \cite{shao2024deepseekmathpushinglimitsmathematical} samples \(G\) completions and computes their rewards \(\{r_i\}_{i=1}^{G}\). Define the $t^{\text{th}}$ token of the $i^{\text{th}}$ completion as \(o_{i,t}\). GRPO then assigns a  advantage, \(\hat{A}_{i,t}\) to it.
The optimization objective is defined as: 

\begin{equation}
\begin{split}
 \mathcal{L}_{\text{GRPO}}(\theta) = -\frac{1}{\sum_{i=1}^{G} |o_i|} \sum_{i=1}^{G} \sum_{t=1}^{|o_i|} \biggl[ \min \biggl( w_{i,t} \hat{A}_{i,t}, 
 \text{clip}\left( w_{i,t}, 1-\epsilon, 1+\epsilon \right) \hat{A}_{i,t} \biggr) - \beta \mathcal{D}_{\text{KL}}(\pi_\theta \| \pi_{\text{ref}}) \biggr],
\end{split}
\end{equation}
where the importance weight \(w_{i,t}\) is given by 

\begin{equation}
\label{eq:weight_definition}
w_{i,t} = \frac{\pi_\theta(o_{i,t} \mid q, o_{i,<t})}{[\pi_{\theta_{\text{old}}}(o_{i,t} \mid q, o_{i,<t})]}.
\end{equation}

The clipping threshold \(\epsilon\) controls update magnitude, \([\,\cdot\,]_{\text{old}}\) indicates policy at last step, $\beta$ is the influence of KL divergence $\mathcal{D}_{\text{KL}}$, whose detailed definition can be found in the section \textbf{Detailed equations} in Appendix. 

\paragraph{Limitations of GRPO for small-size LLMs.}
Despite the success of GRPO in large language models (LLMs), small-size LLMs (SLMs) face significant challenges when confronted with complex problems requiring long chains of thought \cite{zhang2025r1}. Due to their inherently limited capacity, SLMs struggle to generate high-quality, reward-worthy candidates for such tasks \cite{li2025adaptive,zheng2025actpaysefficientreinforcement}.
As shown in Figure~\ref{fig:heatmap_grpo_side}, where Qwen3-1.7B is implemented on a code task and it fails to generate correct answers for most queries.
This limitation substantially reduces the probability of sampling high-reward candidates, resulting in advantage signals vanishing (Figure ~\ref{fig:acc1-adv}), thereby constraining the potential performance gains achievable through GRPO in SLMs.

\section{Methodology}

To address the limitations of GRPO on SLMs, we propose incorporating guidance mechanisms into the thinking trajectories, thereby facilitating the sampling of high-quality candidates. We then conduct a comprehensive investigation into various design choices for guidance strategies. Finally, we introduce the \algopt algorithm, which integrates our empirical observations and significantly reduces the need for extensive hyperparameter tuning.

\subsection{Guided GRPO as a Solution}
\begin{figure}[t]
    \centering
    \begin{subfigure}{0.45\textwidth}
        \centering
        \includegraphics[width=\textwidth]{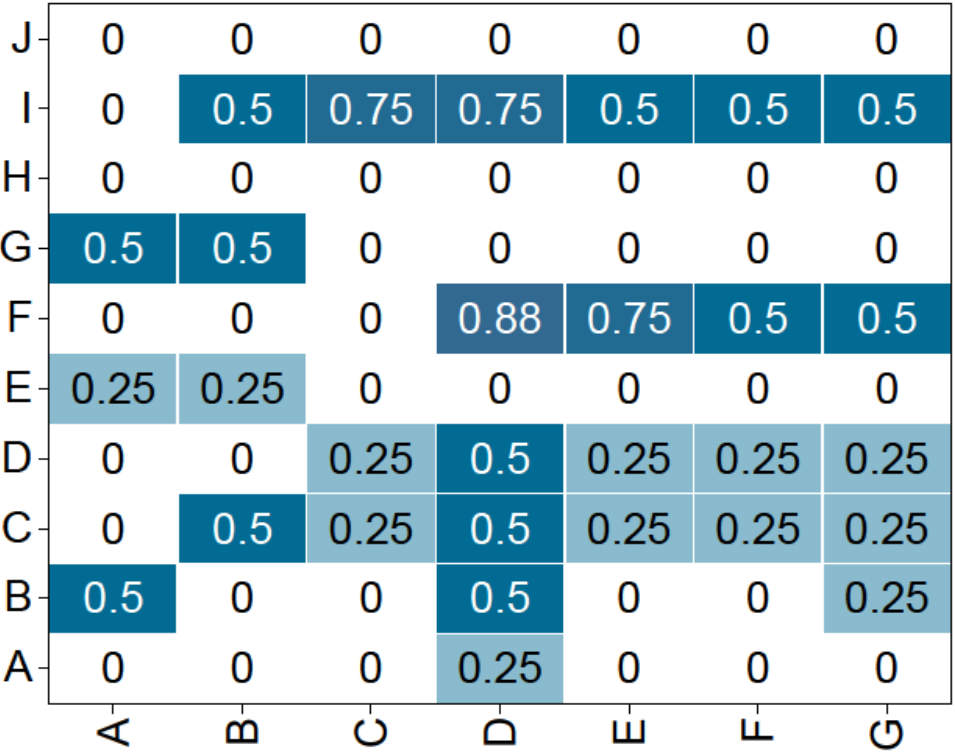}
        \caption{GRPO}
        \label{fig:heatmap_grpo_side}
    \end{subfigure}
    \begin{subfigure}{0.45\textwidth}
        \centering
        \includegraphics[width=\textwidth]{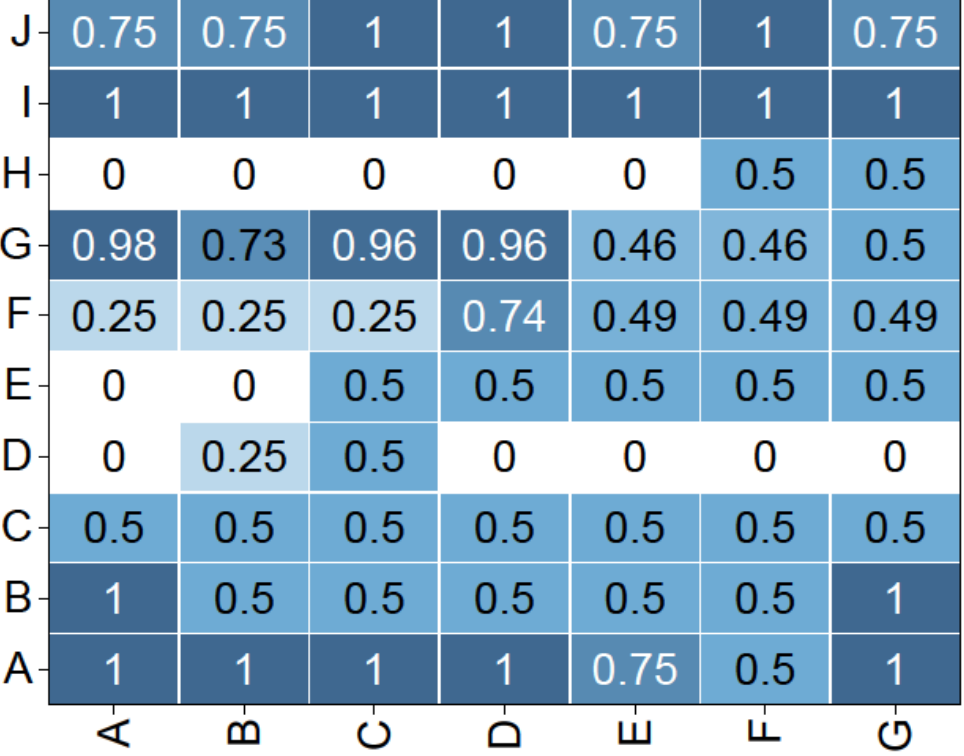}
        \caption{Guided GRPO}
        \label{fig:heatmap_grpo_guided_side}
    \end{subfigure}

    \caption{\textbf{Reward of Guided GRPO.} We fine-tuned Qwen3-1.7B on coding tasks, using 10 roll-outs and generating 280 candidates per batch. The candidates' rewards form a 20x14 matrix. We then applied 2x2 average pooling, reducing it to a 10x7 matrix for clearer visualization. The results demonstrate that when configured with an optimal guidance ratio, G$^2$RPO-A enables the model to sample candidates that yield a significantly denser reward signal.}
    \label{fig:heatmaps_main_horizontal}
\end{figure}

The Guided GRPO can be formulated as:
\begin{align}
    \label{eq:adaguigrpoloss}
    &\mathcal{L}_{\text{guided}}(\theta) = \mathbb{E}_{(q, a) \sim \mathcal{D}, \{g_{i}\}_{i=1}^{G} \sim \mathcal{G}, \{o_{i}\}_{i=1}^{G} \sim \pi_{\text{ref}}(\cdot | q, g_i)}
    \nonumber \\ 
    & \biggl[ \frac{1}{G}  \sum_{i=1}^{G} \frac{1}{|o_i| + |g_i|} \nonumber \biggl(  \sum_{t=1}^{|g_i|}  \min \bigl( w_{g,i,t} \hat{A}_{i,t}, \text{clip}\bigr)\,\hat{A}_{i,t} 
    \nonumber 
     +\sum_{t=1}^{|o_i|} \min \bigl( w_{o,i,t} \hat{A}_{i,t}, \text{clip}\,\bigr) \hat{A}_{i,t} \nonumber -  \beta \mathcal{D}_{\text{KL}}(\pi_\theta \| \pi_{\text{ref}})  \biggr) \biggr],
\end{align}
where $w_{g, i,t}$ and $w_{o, i,t}$ denote the token-level weighting coefficient of guidance $g_i$ and the model outputs $o_i$, respectively.
As shown in Figure~\ref{fig:heatmap_grpo_guided_side}, this guidance enables SLMs to generate higher-reward candidates, potentially overcoming their inherent limitations.

\begin{figure}[t]
    \centering
    \begin{subfigure}{0.45\textwidth}
        \includegraphics[width=\textwidth]{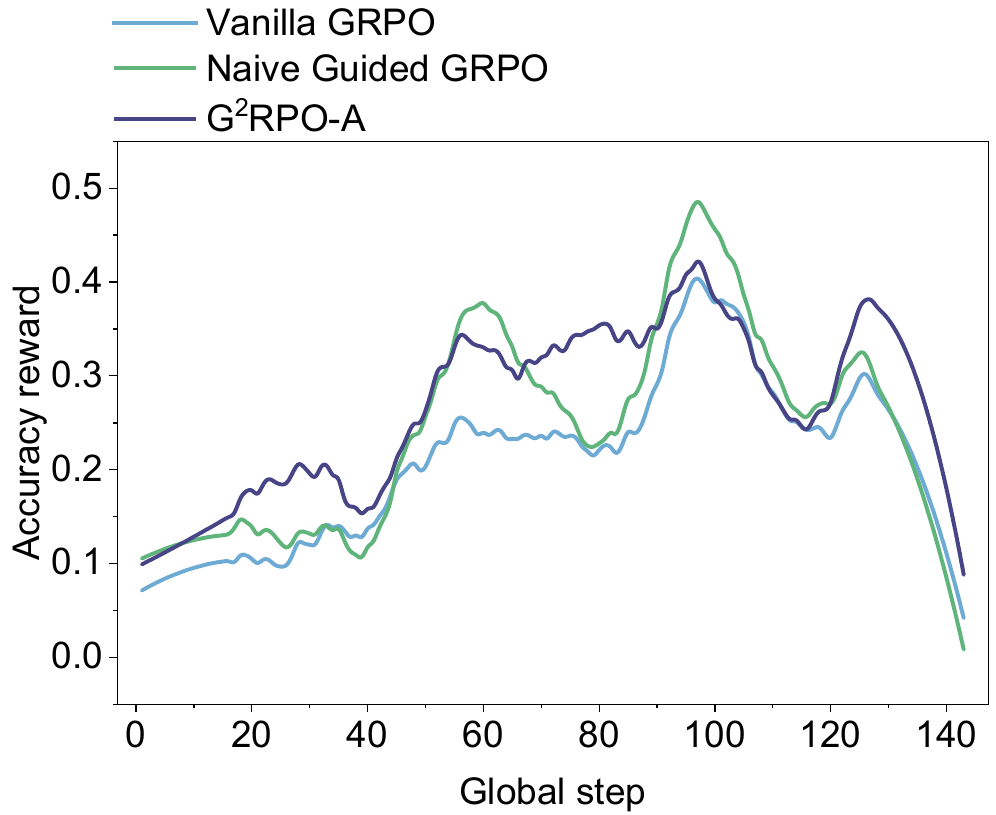}
        \caption{Accuracy Reward}
        \label{fig:acc1}
    \end{subfigure}
    \begin{subfigure}{0.45\textwidth}
        \includegraphics[width=\textwidth]{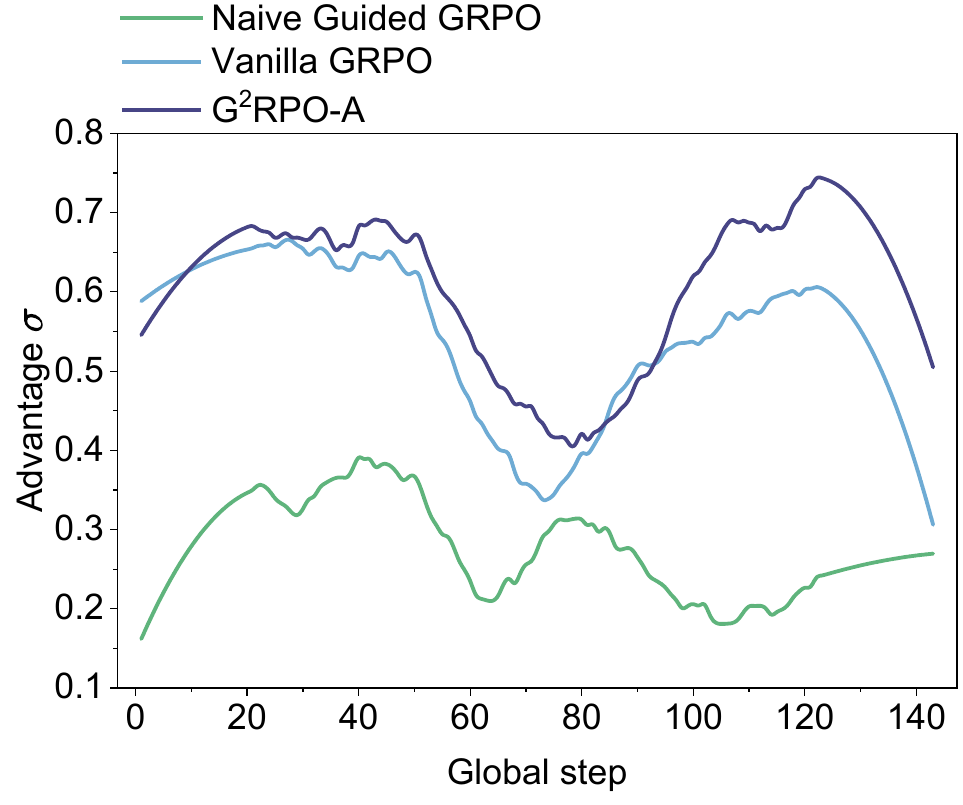}
        \caption{Advantage $\sigma$}
        \label{fig:acc1-adv}
    \end{subfigure}
    \caption{\textbf{Pitfalls of naive Guided GRPO.} We trained Qwen3-1.7B-Base on a curriculum-ordered subset of Math-220K \cite{wang2024openropensourceframework}: problems are presented from easy to hard. Because the curriculum continually increases task difficulty, the accuracy reward does not plateau at a high level--an expected outcome of the CL schedule. This training dynamic indicates that the advantage standard deviation is extremely low under the naive guidance condition, a situation that negatively impacts training efficiency for SLMs.}
    \label{fig:5}
\end{figure}

\paragraph{Naive Guided GRPO fails to boost the final performance.}
Despite increasing expected rewards (Figure~\ref{fig:heatmap_grpo_guided_side}), we found that
\begin{center}
    \emph{simply adding guidance to thinking trajectories of all candidates doesn't enhance final performance and suffers from low advantage. }
\end{center}
As shown in Figure~\ref{fig:acc1} and \ref{fig:acc1-adv}, we train Qwen-3-1.7B-Base on a math dataset sourced from math 220k dataset \cite{wang2024openropensourceframework}, and find that:
(1) Guided GRPO's accuracy reward curve almost matches original GRPO. 
(2) Guided GRPO suffers from low advantage standard deviation, hindering the optimization of the models.
As a result, further investigation is needed to leverage Guided GRPO's higher rewards while ensuring effective training, as the naive approach fails to utilize its potential benefits.

\begin{table}[t]
\centering
\setlength{\tabcolsep}{3pt}
\begin{tabular}{lcccc}
\toprule
MATH500 & $\alpha = \frac{5}{6}$ & $\alpha = \frac{3}{6}$ & $\alpha = \frac{1}{6}$ & $\alpha = 1$ \\
\midrule
$\ell=50$ & $66.80$ & $66.00$ & $67.20$ &$65.10$ \\
$\ell=100$ & $65.20$ & $63.00$ & $66.20$ &$64.70$ \\
$\ell=200$ & $57.60$ & $52.40$ & $62.00$ &$59.30$ \\
$\ell=500$ & $57.80$ & $62.00$ & $68.20$ &$55.80$ \\
\midrule
$\ell=0$ & \multicolumn{4}{c}{62.00} \\
\bottomrule
\end{tabular}
\caption{\textbf{Empirical study on guidance length $\ell$ and guidance ratio $\alpha$.} We use the Qwen2.5-Math-7B as the backbone.}
\label{tab:differentpartialratiodifferentguidancelength}
\end{table}

\subsection{Optimizing Guided GRPO Design}

In this section, we thoroughly examine optimal design choices for Guided GRPO, focusing on guidance ratio of GRPO candidate groups and adjusting guidance strength at different training stages. These investigations aim to maximize the effectiveness of the Guided GRPO and overcome the limitations observed in the naive implementation.

\paragraph{Inner-Group Varied Guidance Ratio.}
The insufficiency of naive guidance suggests that a more nuanced approach is required. We begin by investigating the impact of the guidance ratio $\alpha$.
In each GRPO group of size $G$, we steer only an $\alpha$-fraction of the candidates. Let $g_i$ denote the guidance for the $i$-th candidate (ordered arbitrarily), we have:
\begin{equation}
|g_i| = 0 \quad (i > \alpha G),
\qquad
|g_i| = l \quad (i \le \alpha G).
\end{equation}
That is, the first $\alpha G$ candidates have guidance, while the remaining $(1 - \alpha) G$ candidates evolve freely.
We conduct experiments on the Qwen2.5-Math-7B model~\cite{yang2024qwen2} with a roll-out number $n=6$, training for one epoch on the s1k-1.1 dataset~\cite{muennighoff2025s1simpletesttimescaling}.
We set $\alpha \in \{1/6,\ldots,1\}$ and $l \in \{50,100,\dots,500\}$ tokens, with all accuracies reported on the Math 500 benchmark. The results in Table~\ref{tab:differentpartialratiodifferentguidancelength} show that: 
\begin{itemize}
    \item \textit{Partial inner-group guidance improves model performance.} In most settings, Guided GRPO with guidance provided to only a subset of candidates outperforms the vanilla GRPO, confirming the usefulness of the guidance mechanism.
    \item \textit{For Qwen2.5-Math-7B on the Math500 benchmark, the lowest guidance ratio $\alpha$ combined with a longest guidance window $\ell$ yields the best results.} This suggests that Qwen2.5-Math-7B benefits from infrequent but heavyweight guidance.
\end{itemize}
In summary, selective guidance--directing only few candidates by a long guidance--strikes the best balance between exploration and control, thereby improving model performance. 
Moreover, the optimal guidance ratio varies with both the task domain and model capacity. As Table \ref{tab:adagui-ratio-code}, \ref{tab:adagui-ratio-math-final} shows, smaller models and coding tasks benefit from stronger intervention, whereas larger models and math tasks achieve better results with lighter guidance.

\begin{table}[t]
\centering
\small
\begin{tabular}{c ccc}
\toprule
\textbf{Guidance length} & \multicolumn{3}{c}{\textbf{Decay Policies}} \\
\cmidrule(lr){2-4}
\textbf{Guidance ratio} & \textbf{Step} & \textbf{Linear} & \textbf{Concave} \\
\midrule
$\ell=50$ , $\alpha=0.8333$ & 63.80 & 58.40 & 57.60 \\
$\ell=100$ , $\alpha=0.1667$ & 60.60 & 62.00 & 66.20 \\
$\ell=200$ , $\alpha=0.8333$ & 66.60 & 54.20 & 58.40 \\
$\ell=200$ , $\alpha=0.1667$ & 61.20 & 64.20 & 59.60 \\
$\ell=500$ , $\alpha=0.1667$ & 59.60 & 69.80 & 62.40 \\
\midrule
$\ell=0$ & \multicolumn{3}{c}{62.00} \\
\bottomrule
\end{tabular}
\caption{\textbf{Performance of Guided GRPO under different guidance-length adjustment policies.} We train Qwen2.5-Math-7B and evaluate it on the MATH 500 benchmark. For each guidance-length schedule, we report the results obtained with the guidance ratio that achieves the highest score in Table \ref{tab:differentpartialratiodifferentguidancelength}.}
\label{tab:decay-policies-grouped}
\end{table}

\begin{table*}[t]
\centering
\small
\begin{tabular}{lllccccc}
\toprule
\textbf{Base Model} & \textbf{$\alpha$} & \textbf{Benchmark} & \textbf{Base} & \textbf{GRPO} & \textbf{SFT} & \textbf{G$^2$RPO-A}\\
\midrule
\multirow{3}{*}{\parbox{2.5cm}{Qwen3-0.6B-Base}} & 
\multirow{3}{*}{0.75} & MATH500 & 40.18 & \textbf{54.26} & 50.53 & 51.77 & \\
& & Minerva & 11.43 & 9.57  & 10.40 & \textbf{12.29} & \\
& & gpqa    & 25.49 & 24.51 & 25.49 & \textbf{30.39} & \\
\midrule
\multirow{3}{*}{\parbox{2.5cm}{Qwen3-1.7B-Base}} & 
\multirow{3}{*}{0.25} & MATH500 & 50.96 & 63.74 & 62.11 & \textbf{67.21} & \\
& & Minerva & 13.84 & 16.19 & \textbf{18.89} & 15.10 & \\
& & gpqa    & 27.45 & 29.41 & 24.51 & \textbf{32.35} & \\
\midrule
\multirow{3}{*}{\parbox{2.5cm}{Qwen3-8B-Base}} & 
\multirow{3}{*}{0.14} & MATH500 & 71.32 & 79.49 & 80.29 & \textbf{82.08} & \\
& & Minerva & 33.24 & \textbf{37.51} & 36.60 & 36.42 & \\
& & gpqa    & 43.17 & 44.13 & 42.85 & \textbf{49.72} & \\
\bottomrule
\end{tabular}
\caption{\textbf{Performance of \algopt on Math Tasks.} We report accuracy (\%) on various benchmarks. Models are trained for 5 epochs, and guidance ratios are selected based on the best settings obtained from Table~\ref{tab:adagui-ratio-math-final}.}
\label{tab:benchmark-results-math}
\end{table*}

\begin{table}[t]
\centering
\footnotesize
\setlength{\tabcolsep}{4pt}
\begin{tabular}{lllccc}
\toprule
\textbf{Base Model} & \textbf{$\alpha$} & \textbf{Benchmark} & \textbf{Base} & \textbf{GRPO} & \textbf{G$^2$RPO-A}\\
\midrule
\multirow{4}{*}{Qwen3-0.6B} & 
\multirow{4}{*}{0.75} & MATH500 & 76.20 & 85.37 & \textbf{87.15} \\
& & Minerva & 12.32 & 20.59 & \textbf{21.57} \\
& & gpqa    & 24.51 & 25.45 & \textbf{26.43} \\
& & AIME24    & \textbf{10.00} & 6.67 & \textbf{10.00} \\
& & AIME25    & 13.33 & 20.00 & \textbf{23.33} \\
\midrule
\multirow{4}{*}{Qwen3-1.7B} & 
\multirow{4}{*}{0.25} & MATH500 & 92.71 & \textbf{94.52} & 91.69 \\
& & Minerva & 33.16 & 35.38  & \textbf{38.26} \\
& & gpqa    & 48.23 & 51.68 & \textbf{55.27} \\
& & AIME24    & 46.67 & 56.67 & \textbf{63.33} \\
& & AIME25    & 36.67 & 50.00 & \textbf{53.33} \\
\bottomrule
\end{tabular}
\caption{\textbf{Performance of \algopt on Math Tasks.} The experiment settings are the same with Table \ref{tab:benchmark-results-math}. However, we use extra benchmarks like AIME24 and AIME25 here due to the stronger model performances.}
\label{tab:benchmark-results-math-single-col}
\end{table}

\paragraph{Time Varied Guidance Length.}
Apart from the guidance ratio, Table \ref{tab:differentpartialratiodifferentguidancelength} shows that performance also depends on the guidance length $\ell$. To investigate this further, we evaluate guided GRPO by varying the guidance length during training under three strategies. Those are: 
\begin{equation}
\begin{aligned}
\text{Concave decay:}\quad 
\ell_t &= \ell_0 \Bigl(1 - \tfrac{t}{T}\Bigr)^{\beta} \, \\
\text{Linear decay:}\quad 
\ell_t &= \ell_0 \Bigl(1 - \tfrac{t}{T}\Bigr) \,\\
\text{Stepwise decay:}\quad 
\ell_t &= \ell_0 \,\gamma^{\lfloor t / s\rfloor} \,,
\end{aligned}
\end{equation}
where $T$ is the total training steps, and $\ell_0$ is the initial guidance length. 
The parameter $\beta \in (1, \infty]$ controls the concavity, and $\gamma\in(0,1)$ sets the decay rate, and $s$ specifies the decay interval.

We use the same experiment setting as in Table~\ref{tab:differentpartialratiodifferentguidancelength}, and choose the guidance ratio that performs the best. The results are reported in Table~\ref{tab:decay-policies-grouped}. The results indicate that (1) model quality is highly sensitive to the chosen guidance length $\ell_t$, and (2) no single schedule consistently outperforms the others. \emph{This highlights the need for more effective methods of controlling guidance length.}  

\subsection{\algopt: Sampling Difficulty Motivated Adaptive Guidance}
In this section, we propose an adaptive algorithm that automatically selects the guidance strength $\ell$ at every optimization step. Our approach is inspired by recent work on data filtering and sampling~\citep{bae2025onlinedifficultyfilteringreasoning, xiong2025minimalistapproachllmreasoning, shi2025efficientreinforcementfinetuningadaptive}
, which removes examples that yield uniformly low or uniformly high rewards. Such “uninformative” samples--being either too easy or too hard--contribute little to learning and can even destabilize training. The pseudo-code can be found in Appendix.

\paragraph{Guidance length adjustment.} 
Our key idea is to control the difficulty of training samples by dynamically adjusting the guidance length, taking into account the ongoing training states.
At each training step $k$, the guidance $\ell_{k+1}$ is determined by the following equation: 
\begin{equation}
    \ell_{k+1} = \ell_k \cdot \frac{\text{min}(\mathcal{T}, k) r_{k}}{\sum_{\tau=1}^{\text{min}(\mathcal{T}, k)} r_{k-\tau}},
    \label{equ:adapt-guidance-length}
\end{equation}
where $r_k$ is the average reward of the $k$-th training step, $\mathcal{T}$ is a hyperparameter that controls the number of history steps we considered, and we found that setting $\mathcal{T}=2$ is already sufficient for noticeably improving Guided GRPO performance (Table~\ref{tab:guidance_ablation_math}, \ref{tab:guidance_ablation_code}).

Equation \ref{equ:adapt-guidance-length} implies the following dynamics:
\begin{itemize}
    \item When recent rewards rise, $\ell_{k}$ is reduced, making the next batch of examples harder.
    \item When recent rewards fall, $\ell_{k}$ is increased, making the next batch easier.
\end{itemize}
Thus, the training difficulty is automatically and continuously adjusted to match the model’s current competence.

\begin{table*}[t]
\centering
\small
\begin{tabular}{llcccccc}
\toprule
\textbf{Base Model} & \textbf{Guidance Ratio} & \textbf{Benchmark} & \textbf{Base Perf.} & \textbf{GRPO} & \textbf{SFT} & \textbf{G$^2$RPO-A} \\
\midrule
\multirow{2}{*}{\parbox{2.5cm}{Qwen3-0.6B}} & 
\multirow{2}{*}{0.75} & humaneval & 32.32 & 38.89 & 40.33 & \textbf{44.96} & \\
& & Live Code bench & 17.07 & 22.22 & 13.58 & \textbf{23.14} & \\
\midrule
\multirow{2}{*}{\parbox{2.5cm}{Qwen3-1.7B}} & 
\multirow{2}{*}{1} & humaneval & 46.08 & 67.65 & 63.34 & \textbf{75.93} & \\
& & Live Code bench & 34.31 & 53.14 & \textbf{56.33} & 51.96 & \\
\midrule
\multirow{2}{*}{\parbox{2.5cm}{Qwen3-8B}} & 
\multirow{2}{*}{0.57} & humaneval & 64.36 & \textbf{81.48} & 77.42 & 80.33 & \\
& & Live Code bench & 60.58 & 77.12 & 63.82 & \textbf{79.71} & \\
\bottomrule
\end{tabular}
\caption{\textbf{Performance of \algopt on Code Tasks.} We report accuracy (\%) on various benchmarks. Models are trained for 5 epochs, and guidance ratios are selected based on the best settings obtained from Table~\ref{tab:adagui-ratio-code}.}
\label{tab:benchmark-results-code}
\end{table*}

\paragraph{Curriculum learning for further improvements.} 
Equation \ref{equ:adapt-guidance-length} shows that the adaptive guidance-length controller updates $\ell$ by comparing the current reward with rewards from previous steps. When consecutive batches differ markedly in difficulty, these reward variations no longer reflect the model’s true learning progress, which in turn degrades \algopt’s performance. 

\begin{table}[t]
\centering
\small
\begin{tabular}{l cccc}
\toprule
& \multicolumn{2}{c}{\textbf{Random}} & \multicolumn{2}{c}{\textbf{CL}} \\
\cmidrule(lr){2-3} \cmidrule(lr){4-5}
& \textbf{GRPO} & \textbf{G$^2$RPO-A} & \textbf{GRPO} & \textbf{G$^2$RPO-A} \\
\midrule
\multicolumn{5}{l}{\textbf{Qwen3-1.7-Base}} \\
\cmidrule(r){1-1}
MATH500 & 53.81 & 57.67 & 52.05 & \textbf{58.94} \\
Minarva  & 12.41 & 15.12 & 14.98 & \textbf{16.69} \\
gpqa     & 24.79 & 23.53 & \textbf{27.45} & 25.49 \\
\midrule
\multicolumn{5}{l}{\textbf{Qwen3-0.6B-Base}} \\
\cmidrule(r){1-1}
Math 500 & 43.25 & 50.72 & 48.16 & \textbf{53.59} \\
Minarva  & 11.04 & \textbf{11.21} & 9.66  & 10.08 \\
gpqa     & 23.1  & 25.49 & 24.51 & \textbf{32.35} \\
\bottomrule
\end{tabular}
\caption{\textbf{Comparison of training with random order and curriculum learning (CL) order} across different models and benchmarks.}
\label{tab:training_order_comparison}
\end{table}

\begin{table}[!t]
\centering
\small
\begin{tabular}{lccccc}
\toprule
\textbf{Setting} & 
\textbf{Level 1} & 
\textbf{Level 2} & 
\textbf{Level 3} & 
\textbf{Level 4} & 
\textbf{Level 5} \\
\midrule
Remove & 76.74 & 71.11 & 50.00 & 35.00 & 24.00 \\
Replace & \textbf{88.37} & 75.56 & 54.00 & 37.00 & 18.00 \\
Original & 86.05 & \textbf{77.77} & \textbf{60.00} & \textbf{43.00} & \textbf{28.00} \\
\bottomrule
\end{tabular}
\caption{\textbf{Performance of GRPO with different sample-filtering methods.} We train Qwen3-1.7B model using \algopt, with $\alpha=0.25$. In the \textsc{Remove} setting all hard samples are excluded from the original dataset, whereas in the \textsc{Replace} setting each hard sample is substituted with a sample of moderate difficulty.}
\label{tab:dataset-arrangement-results}
\end{table}

To eliminate this mismatch, we embed a curriculum-learning (CL) strategy \cite{parashar2025curriculumreinforcementlearningeasy,shi2025efficientreinforcementfinetuningadaptive,zhou2025discobalancesscalesadaptive}. Concretely, we sort the samples by difficulty. Using math task as an example, we rank examples by source, yielding five ascending difficulty tiers: cn\_contest, aops\_forum, amc\_aime, olympiads, and olympiads\_ref. We also tested ADARFT \cite{shi2025efficientreinforcementfinetuningadaptive}, which orders samples by success rate, but its buckets proved uninformative in our cases—most questions were either trivial or impossible (see Appendix Figure 6)—so it failed to separate difficulty levels effectively. Table~\ref{tab:training_order_comparison} shows that both the performance of vanilla GRPO and \algopt boosted by CL.

\paragraph{Compare \algopt to sample-filtering methods.}
Earlier work argues that policy-gradient training benefits most from mid-level queries. \citet{bae2025onlinedifficultyfilteringreasoning} keep only moderate-difficulty batches via an online filter, and Reinforce-Rej~\cite{xiong2025minimalistapproachllmreasoning} discards both the easiest and hardest examples to preserve batch quality.
Our experiments show that this exclusion is counter-productive: removing hard problems deprives the model of vital learning signals and lowers accuracy on challenging tasks. Table~\ref{tab:dataset-arrangement-results} confirms that either dropping hard items or substituting them with moderate ones reduces Level 4 and 5 test accuracy.
\algopt avoids this pitfall by retaining tough examples and attaching adaptive guidance to them, thus exploiting the full difficulty spectrum without sacrificing performance.

\section{Experiments}

\begin{table}[t]
\centering
\small
\begin{tabular}{lccccc}
\toprule
Qwen3-1.7B & $\alpha=0$ & $\alpha=\frac{1}{4}$ & $\alpha=\frac{2}{4}$ & $\alpha=\frac{3}{4}$ & $\alpha=1$ \\
\midrule
humaneval       & 68.52 & 59.88 & 64.81 & \textbf{72.22} & 70.81 \\
LCB & 28.43 & 19.61 & 23.53 & 30.39 & \textbf{35.72} \\
\midrule[\heavyrulewidth]
Qwen3-0.6B & $\alpha=0$ & $\alpha=\frac{1}{4}$ & $\alpha=\frac{2}{4}$ & $\alpha=\frac{3}{4}$ & $\alpha=1$  \\
\midrule
humaneval       & 41.98 & 32.10 & 27.72 & 38.89 & \textbf{49.38} \\
LCB & 12.75 & 11.76  & 9.80  & \textbf{18.63}  & 12.75 \\
\bottomrule
\end{tabular}
\caption{\textbf{Ablation studies on guidance ratio $\alpha$ for Code Tasks.} The group size is set to 12. The initial guidance length for G$^2$RPO-A is set to 3072. The LCB indicates Live Code Bench.}
\label{tab:adagui-ratio-code}
\end{table}

\begin{table}[t]
\centering
\small
\begin{tabular}{lccccc}
\toprule
& $\alpha=0$ & $\alpha=\frac{1}{4}$ & $\alpha=\frac{2}{4}$ & $\alpha=\frac{3}{4}$ & $\alpha=1$ \\
\midrule
\multicolumn{6}{l}{\textbf{Qwen3-1.7B-Base}} \\
\cmidrule(r){1-1} 
MATH500 & 52.05 & \textbf{58.71} & 53.09 & 55.53 & 45.95 \\
Minerva & 14.98 & 16.69 & 16.25 & \textbf{18.21} & 16.11 \\
gpqa    & 27.45 & 25.49 & \textbf{30.39} & 25.49 & 22.55 \\
\midrule
\multicolumn{6}{l}{\textbf{Qwen3-0.6B-Base}} \\
\cmidrule(r){1-1}
MATH500 & 48.16 & 49.59 & 50.94 & \textbf{53.50} & 38.42 \\
Minerva & 9.66  & 9.10  & 8.96  & 10.08 & \textbf{15.69} \\
gpqa    & 24.51 & 19.61 & 31.37 & \textbf{32.35} & 25.49 \\
\bottomrule
\end{tabular}
\caption{\textbf{Ablation studies on guidance ratio $\alpha$ for Math Tasks.} The group size is set to 12. The initial guidance length for G$^2$RPO-A is set to 3072.}
\label{tab:adagui-ratio-math-final}
\end{table}

\begin{table}[t]
\centering
\footnotesize
\setlength{\tabcolsep}{3pt} 
\begin{tabular}{lcccccc}
\toprule
& \multirow{2}{*}{\textbf{GRPO}} & \multicolumn{3}{c}{\textbf{Fixed Guidance}} & \multirow{2}{*}{\textbf{{RDP}}} & \multirow{2}{*}{\textbf{G$^2$RPO-A}} \\
\cmidrule(lr){3-5} 
& & \textbf{3072} & \textbf{2048} & \textbf{1024} & & \\
\midrule
\multicolumn{7}{l}{\textbf{Qwen3-1.7B-Base}} \\
\cmidrule(r){1-1} 
MATH500 & 52.05 & 51.28 & 60.52 & 46.78 & 51.02 & \textbf{58.71} \\
Minerva & 14.98 & 14.40 & 17.16 & 12.22 & 17.99 & \textbf{22.46} \\
gpqa    & \textbf{27.45} & 25.00 & 24.51 & 23.53 & 22.13 & 25.49 \\
\midrule
\multicolumn{7}{l}{\textbf{Qwen3-0.6B-Base}} \\
\cmidrule(r){1-1}
MATH500 & 48.16 & 55.80 & 54.17 & 52.69 & \textbf{55.97} & 53.50 \\
Minerva & 9.66  & 13.27 & 15.26 & 11.78 & 14.32 & \textbf{15.69} \\
gpqa    & 24.51 & 24.00 & 21.57 & 22.55 & 26.00 & \textbf{32.35} \\
\bottomrule
\end{tabular}
\caption{\textbf{Guidance-length ablation on Math Tasks.} Each run uses the optimal guidance ratio reported in Table \ref{tab:adagui-ratio-math-final}. The initial guidance budget for \algopt is fixed at 3,072 tokens. RDP refers to the rule-based decay policy.}
\label{tab:guidance_ablation_math}
\end{table}

\subsection{Experiment Settings}

In this section, we outline the experiment settings we used, and more details about dataset filtering methods and evaluation on more models can be found in Appendix.

\paragraph{Datasets and models.} 
We conduct experiments on math and code tasks. In detail, 
\begin{itemize}
    \item \textbf{Mathematical reasoning tasks.} We construct a clean subset of the Open-R1 math-220k corpus \cite{wang2024openropensourceframework}. Problems are kept only if their solution trajectories are (i) complete, (ii) factually correct, and (iii) syntactically parsable.
    \item \textbf{Code generation.} For programming experiments we adopt the Verifiable-Coding-Problems-Python benchmark from Open-R1. For every problem we automatically generate a chain-of-thought with QWQ-32B-preview~\cite{team2024qwq}. These traces are later used as guidance by our proposed \algopt training procedure.
\end{itemize}
We use Qwen3 series \cite{yang2025qwen3} for both tasks. Results of DeepSeek-Math-7B-Base \cite{shao2024deepseekmathpushinglimitsmathematical} for math and DeepSeek-Coder-6.7B-Base \cite{guo2024deepseek} for code also included in Appendix. Unless specifically mentioned, CL is used for all experiments for fair comparison, and we also conducted ablation studies in Table~\ref{tab:training_order_comparison}.

\paragraph{Evaluation protocol.}
We assess our models mainly on three public mathematical–reasoning benchmarks—\textsc{Math500} \cite{hendrycks2021measuring}, \textsc{Minerva-Math} \cite{lewkowycz2022solving}, and \textsc{GPQA} \cite{rein2024gpqa}. For the mathematical training of Qwen3-1.7B and Qwen3-0.6B, \textsc{AIME24} \cite{numina_math_datasets} and \textsc{AIME25} benchmarks are also used. And for code tasks, we use two benchmarks: \textsc{humaneval} \cite{chen2021evaluatinglargelanguagemodels} and \textsc{Live Code Bench} \cite{jain2024livecodebench}. Decoding hyper-parameters are fixed to: temperature $=0.6$, $\mathrm{top}\text{-}p = 0.95$, and $\mathrm{top}\text{-}k = 20$.  
Unless otherwise noted, we generate with a batch size of $128$ and permit a token budget between $1{,}024$ and $25{,}000$, based on each model’s context window.

\paragraph{Training details.}
Our G$^{2}$RPO-A algorithm is implemented on top of the fully open-source \textsc{Open-R1} framework \cite{openr1}.  
We use the following hyper-parameters: 
(i) number of roll-outs per sample set to $12$ for 0.6B and 1.7B backbones, and $7$ for 7B and 8B backbones;  
(ii) initial learning rate $1\times10^{-6}$, decayed with a cosine schedule and a warm-up ratio of $0.1$;  
(iii) a training set of $1{,}000$ problems for $5$ epochs. Note that for ablation experiments, only 1 epoch is implemented in our training. (iv) All models are trained on 8 A100 GPUs. 

\subsection{Numerical Results}

\paragraph{Superior performance of \algopt.}
As reported in Table~\ref{tab:benchmark-results-math},~\ref{tab:benchmark-results-math-single-col}, and~\ref{tab:benchmark-results-code},
(i) our proposed \algopt\ markedly surpasses vanilla GRPO on nearly every benchmark, and
(ii) all RL‐based methods outperform both the frozen base checkpoints and their SFT variants, mirroring trends previously observed in the literature.

\paragraph{Effect of the guidance ratio $\boldsymbol{\alpha}$.}  
Table~\ref{tab:adagui-ratio-code}, \ref{tab:adagui-ratio-math-final} show that 
(1) larger models benefit from weaker guidance—e.g., Qwen3-1.7B peaks at $\alpha{=}0.25/0.5$ on Math, whereas the smaller Qwen3-0.6B prefers $\alpha{=}0.75$;  
(2) Code tasks consistently require a higher guidance ratio than Math.

\paragraph{Ablation on guidance–length schedules.}
Table~\ref{tab:guidance_ablation_math}, \ref{tab:guidance_ablation_code} contrast our adaptive scheme (\algopt) with (i) fixed guidance and (ii) a rule-based decay policy (RDP).  
(1) \algopt achieves the best score on almost every model–benchmark pair, confirming the benefit of on-the-fly adjustment.  
(2) For fixed guidance, the optimal value varies across both tasks and model sizes, with no clear global pattern, underscoring the need for an adaptive mechanism such as \algopt.

\begin{table}[!t]
\centering
\footnotesize
\setlength{\tabcolsep}{3pt} 
\begin{tabular}{lcccccc}
\toprule
& \multirow{2}{*}{\textbf{GRPO}} & \multicolumn{3}{c}{\textbf{Fixed Guidance}} & \multirow{2}{*}{\textbf{{RDP}}} & \multirow{2}{*}{\textbf{G$^2$RPO-A}} \\
\cmidrule(lr){3-5} 
& & \textbf{3072} & \textbf{2048} & \textbf{1024} & & \\
\midrule
\multicolumn{7}{l}{\textbf{Qwen3-1.7B}} \\
\cmidrule(r){1-1} 
humaneval & 68.52 & 58.64 & 58.02 & 60.49 & 69.29 & \textbf{70.81} \\
LCB       & 23.53 & 29.41 & 28.43 & 31.37 & 26.47 & \textbf{35.72} \\
\midrule
\multicolumn{7}{l}{\textbf{Qwen3-0.6B}} \\
\cmidrule(r){1-1}
humaneval & 38.89 & 43.93 & 36.54 & 38.40 & 42.27 & \textbf{49.38} \\
LCB       & 12.75 & \textbf{13.73} & 10.78 & 9.80 & 11.67 & 12.75 \\
\bottomrule
\end{tabular}
\caption{\textbf{Guidance-length ablation on Code Tasks.} Each run uses the optimal guidance ratio reported in Table \ref{tab:adagui-ratio-code}. The initial guidance budget for \algopt is fixed at 3,072 tokens. RDP refers to the rule-based decay policy.}
\label{tab:guidance_ablation_code}
\end{table}

\section{Conclusion and Future Work}

We introduce a method that injects ground-truth guidance into the thinking trajectories produced during GRPO roll-outs, thereby improving the performance of small-scale LLMs. After an extensive study of guidance configurations, we observe that the guidance ratio is significant in guidance mechanism and the optimal guidance length is context-dependent and, based on this, we develop \algopt, an auto-tuned approach. Experiments on mathematical reasoning and code generation demonstrate that \algopt consistently boosts accuracy. In future work, we plan to evaluate \algopt across a broader range of tasks and model architectures, which we believe will further benefit the community.

\bibliographystyle{plainnat}
\bibliography{references}
\end{document}